\documentclass{llncs} 
\usepackage[T1]{fontenc}
\usepackage[utf8]{inputenc}
\usepackage{latexsym,amsmath,amsfonts,amssymb,braket,subcaption}
\usepackage{array}
\usepackage{listings}
\usepackage{cite}
\usepackage{amsfonts}
\usepackage{bbm}
\usepackage{amssymb}
\usepackage{graphicx}
\usepackage{caption}
\usepackage{lmodern}
\usepackage[table,dvipsnames]{xcolor}
\usepackage{booktabs}
\usepackage{siunitx}
\usepackage{enumitem}
\usepackage{mathtools}
\usepackage{makecell}
\usepackage{url}
\usepackage[ruled,linesnumbered]{algorithm2e}
\usepackage[noend]{algpseudocode}
\usepackage[table]{xcolor}
\usepackage[labelfont=bf]{caption}
\usepackage[colorlinks=true,allcolors=blue]{hyperref}
\usepackage{makecell}

\DeclareMathOperator{\CSP}{CSP}
\DeclareMathOperator*{\COP}{COP}

\DeclareMathOperator*{\argmax}{arg\,max}
\DeclareMathOperator*{\argmin}{arg\,min}

\newcommand\Rocs[1]{\textcolor{YellowGreen}{}}
\newcommand\tias[1]{\textcolor{YellowGreen}{}}
\newcommand\maxime[1]{\textcolor{TealBlue}{}}
\newcommand{\jay}[1]{\textcolor{TealBlue}{}}
\begin{document}

\title{Hybrid Classification and Reasoning \\ for Image-based Constraint Solving}

\author{Maxime Mulamba \orcidID{0000-0002-9122-926X}\and  Jayanta Mandi  \orcidID{0000-0001-8675-8178}\and  Rocsildes Canoy  \orcidID{0000-0003-1810-082X} \and Tias Guns\orcidID{0000-0002-2156-2155}}

\institute{Data Analytics Laboratory, Vrije Universiteit Brussel, \{firstname.lastname\}@vub.be}


\maketitle
\begin{abstract}
There is an increased interest in solving complex constrained problems where part of the input is not given as facts, but received as raw sensor data such as images or speech. We will use `visual sudoku' as a prototype problem, where the given cell digits are handwritten and provided as an image thereof. In this case, one first has to train and use a classifier to label the images, so that the labels can be used for solving the problem.
In this paper, we explore the hybridisation of classifying the images with the reasoning of a constraint solver. We show that pure constraint reasoning on predictions does not give satisfactory results. Instead, we explore the possibilities of a tighter integration, by exposing the probabilistic estimates of the classifier to the constraint solver. This allows joint inference on these probabilistic estimates, where we use the solver to find the maximum likelihood solution. We explore the trade-off between the power of the classifier and the power of the constraint reasoning, as well as further integration through the additional use of structural knowledge. Furthermore, 
we investigate the effect of calibration of the probabilistic estimates on the reasoning. Our results show that such hybrid approaches vastly outperform a separate approach, which encourages a further integration of prediction (probabilities) and constraint solving.
\keywords{ Constraint Reasoning,  Visual sudoku,  Joint Inference, Prediction and Optimisation}   
\end{abstract}

\section{Introduction}
\label{intro}

Artificial intelligence (AI) is defined as “systems that display intelligent behaviour by analysing their environment and taking actions – with some degree of autonomy – to achieve specific goals.”\cite{AIHLEG}.
In that regard, recent advancements in deep neural network (DNN) architectures have achieved highly accurate performance in object and speech recognition and classification. However, many real life problems are relational, where inference on one instance is related to another through various constraints and logical reasoning.
Attaining good performance in tasks which require reasoning over constraints and relations still remains elusive. 
The DNN architectures rely heavily on learning latent representation from the training datasets \cite{van2017neural}. The main reason deep architectures struggle in constraint reasoning is that the nuances of the relationship between entities are often lost in the latent representation. For instance, when solving a sudoku, a DNN model would take the partially filled sudoku as an input and would then be expected to produce the solved sudoku as output. In this process, the model fails to comprehend the interactions among different cells.


Moreover, the high quality performance of DNNs at complex tasks comes at a cost. As DNN models fail to comprehend the logical reasoning, they have to adjust to gradual feedback of the error signals. As a consequence, to be proficient in any simple task, a DNN needs an enormous amount of data. As an example, to be an efficient video-gamer, a DNN model has to play a game for more than 900 hours \cite{lake2017building}. 
Motivated by such deficiencies, integrating logical and relational reasoning into DNN architecture has increasingly gained more attention. 

In trying to bridge deep learning and logical reasoning, Wang et al.~\cite{wang2019satnet}  propose SATNet, a differentiable satisfiability solver that can be used to learn both constraints and image classification through backpropagation. Internally, it uses a quadratic SDP relaxation of a MaxSAT model, and hence learns a relaxed representation of the constraints. We argue that in many cases, there is no need to learn everything end-to-end. Indeed, in a visual sudoku setting, while the constraints are easy to specify in a formal language, the image classification task is difficult for a machine to capture. Hence, we seek to bridge deep learning and logical reasoning by directly plugging the (probabilistic) output of the deep learning into a constraint solver that reasons over the relevant hard constraints.

In this work, we present a framework where we perform \textit{joint inference} \cite{punyakanok2004semantic,poon2007joint, riedel2012improving} over the different predictions, by integrating machine learning inference with first and second order logic. Specifically, instead of solving a constraint programming (\textit{CP}) problem over a set of independently predictied values, we use CP to do joint inference over a set of probability vectors. The training of the DNN happens on individual image instances, as is typically done. Effectively, our framework can be considered as a \emph{forward-only layer} on top of the predictions of a pre-trained network.

Specifically, we consider the ``visual sudoku'' problem where images of digits of some cells in the sudoku grid are fed as input. We first predict the digits using a DNN model and then use a CP solver to solve the sudoku puzzle. A conventional approach would use the predictions of the DNN as inputs to the CP. As the DNN model is not aware of the constraints of the sudoku problem, it misses the opportunity to improve its prediction by taking the constraints into account. When the predictions of the DNN are directly fed into the CP solver, in case of any error, the CP model is bound to fail. Note that in this case, even one prediction error will result in the failure of the whole problem.  

We improve the process by considering the predicted class probabilities instead of directly using the \textit{arg max} prediction. The advantage of our approach is that by avoiding hard assignments prior to the CP solver, we enable the CP solver to \textit{correct the errors} of the DNN model. In this way, we use CP to do joint inference, which ensures that the predictions will respect the constraints of the problem. 

The contributions of the paper are as follows:
\begin{itemize}
\item We explore hybridisation of classification and constraint reasoning on the visual sudoku problem;
\item We show that constraint reasoning over the probabilistic predictions outperforms a pure reasoning approach, and that we can further improve by taking higher-order relations into account;
\item We investigate the increased computational cost of reasoning over the probabilities, and the trade-offs possible when limiting the reasoning to the top-k probabilities.
\item We experimentally explore the interaction of predictive power with the power of discrete reasoning, showing correction factors of 10\% and more, as well as the effect of using \textit{calibrated} probabilistic classifiers.
\end{itemize}

\section{Related work}
\subsubsection{Predict-and-optimize}
Our work is closely related to the growing body of research at the intersection of machine learning (ML) and combinatorial optimization \cite{ifrim2012properties,kool2019attention,mukhopadhyay2017prioritized} where the predictions of an ML model is fed into a downstream optimization oracle. 
In most applications, feeding machine learning predictions directly into a combinatorial optimization problem may not be the most suitable approach.
Bengio \cite{bengio1997using} compared two ML approaches for optimizing stock returns---one uses a neural network model for predicting financial prices, and the second model makes use of a task-based loss function. Experimental results show that the second model delivers better optimized return. The results also suggest a closer integration of ML and optimization.

In this regard, Wilder et al. \cite{wilder2019melding} propose a framework which trains the weight of the ML model directly from the task-loss of the downstream combinatorial problem from its continuous relaxation. The end-to-end model of \cite{wang2019satnet} learns the constraints of a satisfiability problem by considering a differentiable SDP relaxation of the problem. A similar work\cite{mandi2019smart} trains an ML model by considering a convex surrogate of the task-loss.
\jay{This reference can be removed in case of space constraints}
 
Our work differs from these as we do not focus on end-to-end learning. Rather, we enhance the predictions of an ML model by using CP to do joint inference over the raw probability vectors. In this way, we are taking the constraint interaction of the combinatorial problem into account.
\subsubsection{Joint inference}
Our work is also aligned with the research in joint inference. For example, Poon and Domingos \cite{poon2007joint} have shown its advantage for information extraction in the context of citation matching. Recent work in linguistic semantic analysis of Wang et al.\cite{wang2019joint} forms a factor graph from the DNN output by encoding it into logical predicates and performs a joint inference over the factor graph. Several other works\cite{li2011joint, li2013joint,chen2018encoding} focus on leveraging joint inference in DNN architecture for relation extraction from natural language. Our work differs from these, as we perform probabilistic inference on combinatorial constraint solving problem where one inference is linked with another by hard constraints. 
\subsubsection{Training with Constraints} Various works have introduced methods to enforce constraints on the outputs of an NN.  
One of the earlier work\cite{platt1988constrained} does this by optimizing the Lagrangian coefficients of the constraints at every parameter update of the network. But this would not be feasible in the context of deep neural network as very large dimension matrices must be numerically solved for each parameter update \cite{marquez2017imposing}. 
Pathak et al. \cite{pathak2015constrained} introduce CCNN for image segmentation with size constraints where they introduce latent probability distributions over the labels and impose constraints on the latent distribution enabling efficient Lagrangian dual optimization. However, one drawback is, this involves solving an optimization problem at each iteration. Márquez-Neila et al.\cite{marquez2017imposing} use a Lagrangian based Krylov subspace approach to enforce linear equality constraints on the output of an NN. But this approach is not found to be scalable to large problem instances. 
The proposed framework of \cite{li2019logic} quantifies inconsistencies of the NN output with respect to  the logic constraints and is able to significantly reduce inconsistent constraint violating outcomes by training the model to minimize inconsistency loss. 

The closest work to ours is \cite{punyakanok2004semantic},
where Punyakanok et al. train a multiclass classifier to identify the label of an argument in the context of semantic role labeling and then feed the prediction scores of each argument to an Integer Linear Programming solver so that the final inferences abide by some predefined linguistic constraints.  
\section{Preliminaries}
\subsubsection{CSP and COP}
The concept of a constraint satisfaction problem (CSP) is fundamental in constraint programming 
\cite{rossi2006handbook}. A CSP is formulated as a triplet $(V,D,C)$, where $V$ is a set of decision  variables, 
each of which has its possible values in a domain contained in the set $D$, and $C$ is a set of constraints that need to be satisfied over the variables in $V$. In most cases, we are not only interested in knowing whether a constrained problem is solvable, but we want the \emph{best} possible solution according to an objective.

A Constraint Optimization Problem $COP(V,D,C,o)$ finds a feasible solution of optimum value with respect to an objective function $o$ over the variables. In case of a minimisation problem, we have: $S \in COP(V,D,C,o)$ iff $S \in CSP(V,D,C)$ and $\nexists T \in CSP(V,D,C)$ with $o(T) < o(S)$.

\subsubsection{Sudoku}
In our work we consider a prototype CSP, namely the sudoku. Sudoku is a number puzzle, played on a partially filled $9$x$9$ grid. The goal is to find the unique solution by filling in the empty grid cells with numbers from $1$ to $9$ in such a way that each row, each column and each of the nine $3$x$3$ subgrids contain all the numbers from $1$ to $9$ once and only once.

Formally, the sudoku is a $CSP(V,D,C)$ where $V$ is the set of variables $v_{ij}$ $(i,j \in \{1,...,9\})$ for every cell  in the grid, and $D(v_{ij}) = \{1,...,9\}$ for each $v_{ij} \in V$.
We separate the sudoku constraints into two parts: the set of constraints $C_{given}$ defining the assignment of numbers in the filled cells (hereinafter referred to as the \emph{givens}) of the grid and the set of constraints $C_{rules}$ defined by the rules of sudoku. 

Formally, $C_{rules}$ consists of the following constraints:
\begin{align}
\begin{split}
\forall i \in \{1,...,9\} \ \quad &\texttt{alldifferent} \{v_{i1},...,v_{i9}\} \\
\forall j \in \{1,...,9\} \ \quad &\texttt{alldifferent} \{v_{1j},...,v_{9j}\} \\
\forall i,j \in \{1,4,7\} \ \quad &\texttt{alldifferent} \{
  v_{ij},..., v_{(i+2)j}, ~ v_{i(j+1)},..., v_{(i+2)(j+1)}, \\
& \quad \quad \quad \quad \quad \quad ~~ v_{i(j+2)},..., v_{(i+2)(j+2)} \} \\
\forall i,j \in \{1,...,9\} \ \quad &v_{ij} \in \{1,...,9\}
\end{split}
\label{eq:sudoku}
\end{align}
For the given cells, $C_{given}$ is simply an assignment: $D(v_{ij}) = y_{ij}, \ \ \forall v_{ij} \in \{v_{ij} \}^{given} \subset V$ where the $\{y_{ij} \}^{given}$ are known.
Because $V$ and $D$ are obvious from the constraints, we will write $CSP(C_{rules} \wedge C_{given})$ or alternatively $ CSP(C_{rules},$ $\{y_{ij} \}^{given})$ to represent a solution of a sudoku specification.

Sudoku has one additional property, namely that for a set of givens, the solution is unique: $S \in CSP(C_{rules} \wedge C_{given}),\ \nexists T  \in CSP(C_{rules} \wedge C_{given})$, with $T \neq S$. 

\subsubsection{ML Classifier}\label{ML}
We will consider the visual sudoku problem, where the given cells are not provided as facts, but each given cell will be an image of a handwritten digit. We will hence first use Machine Learning (ML) to classify what digit each of the images represents.

Given a dataset of size $n$, $\{(X_i,y_i)\}_{i=1}^n$ with $X_i \in \mathbb{R}^d$ (denoting that each element is a feature vector of $d$ real numbers) and $y_i$ the corresponding  class label, the goal of an ML classifier is to learn a function approximator $f_\theta(X_i)$ (with $\theta$ the trainable parameters of the learning function), such that $f_\theta(X_i) \approx y_i$ for all $(X_i,y_i)$ pairs. In case of a probabilistic classifier, the predicted class label is $\hat{y}_i = f_\theta(X_i) = \argmax_{k} P_\theta(y_i = k | X_i)$ with $P_\theta(y_i = k | X_i)$ the predicted probability that $X_i$ belongs to class $k$~\cite{Goodfellow-et-al-2016}.

Formally, the goal of training is to compute
$
\argmin_\theta \mathcal{L}(f_\theta(X_i),y_i),
$
where $\mathcal{L}(. , .)$ is a \textit{loss function} measuring how well the function approximates the target. An example of a loss function for probabilistic classifiers with $C$ possible classes is the \textit{cross-entropy loss}, defined as:
\begin{align}\label{eq:crossentropy}
    \mathcal{L} = - \dfrac{1}{n} \sum_{i=1}^n \sum_{k=1}^C \mathbbm{1}[y_i = k] \log P_\theta(y_i = k | X_i),
\end{align} 
where $\mathbbm{1}[y_i = k]$ is the indicator function having the value $1$ only when $y_i$ has value $k$, i.e., belongs to class $k$.

\section{Visual sudoku and solution methods} \label{section:methodology}
We first introduce the visual sudoku problem as an example of an image-based constraint solving problem, and then propose three different approaches to solving it by combining classification and reasoning.

\subsubsection{Visual sudoku}
In visual sudoku, the given cells of the sudoku are provided as unlabeled images of handwritten digits. We are also given a large dataset of labeled handwritten digits (the MNIST dataset \cite{lecun1998gradient}). It is inspired by an experiment in~\cite{wang2019satnet}, although we consider the case where the constraints are known and can be used for reasoning.

Formally, $ \texttt{VizSudoku}(C_{rules}, \{X_{ij} \}^{given})$ consists of the rules of sudoku (Eq~\ref{eq:sudoku}), and a set of given images $\{X_{ij} \}^{given}$ each one consisting of a pixel representation of the handwritten digit. The goal is to use a classifier $f_\theta$ on $\{X_{ij} \}^{given}$ such that the predicted labels $\{ \hat{y}_{ij} \}^{given} = \{ f_\theta(X_{ij}) | X_{ij} \in \{X_{ij} \}^{given}\}$ lead to the solution of the sudoku, that is: $CSP(C_{rules}, \{\hat{y}_{ij} \}^{given}) = CSP(C_{rules}, \{{y}_{ij} \}^{given})$ with $y_{ij}$ the true labels of the given images if known.



\subsection{Separate classification and reasoning}\label{section:baseline}

\begin{figure}
  \includegraphics[width=\linewidth]{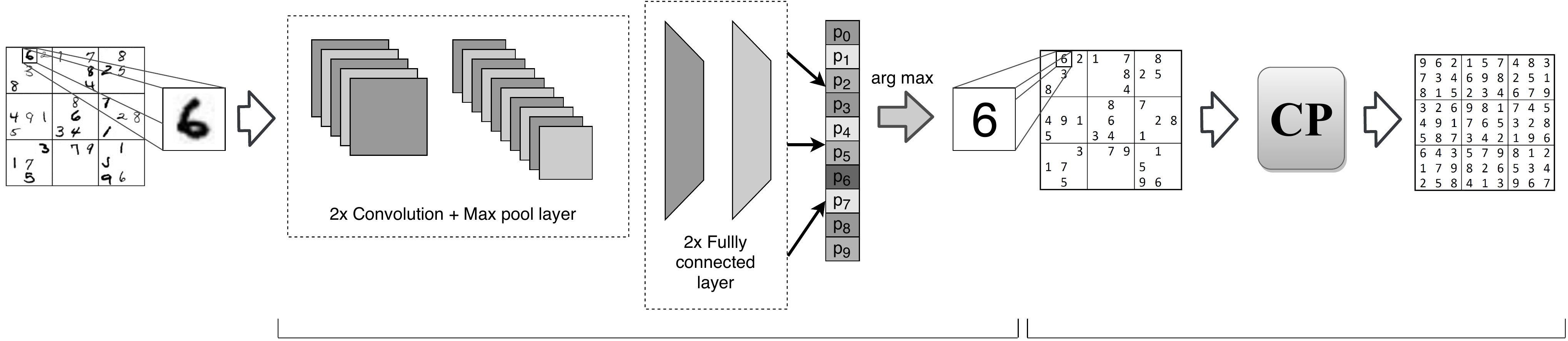}
  \caption{Architecture of separate classification and reasoning approach}
  \label{fig:NeuralNet1}
\end{figure}
The most straightforward approach to solving the visual sudoku problem is to consider the classification and reasoning problems separately. In this approach, first, the most likely digit for each of the given cells are predicted, after which the puzzle is solved using the resulting grid. This will be our baseline approach.  

The baseline approach, explained on Fig.~\ref{fig:NeuralNet1}, is composed of a separate convolutional neural network and a CP solver. The process begins with the training of the DNN on the MNIST training set $\{(X,y)\}$ to obtain a handwritten digit classifier $f_\theta$. Then for each visual sudoku instance, we use the classifier to predict the value of each given cell's image. This takes us from a visual to a purely digital representation of the problem, which is then fed into the CP sudoku solver.
Note, that training is separate from the concept of sudoku, and done on individual images as is standard in image recognition tasks.


Once the model is trained, we use it to solve  $ \texttt{VizSudoku}(C_{rules}, \{X_{ij} \}^{given})$.  
For that, we first predict the digit for each of the given images $\{X_{ij} \}^{given}$.
For each $X_{ij}$ given, the trained DNN computes a class probability for each digit $k$
$P_\theta (y_{ij} = k |X_{ij})$ and predicts the value with the highest probability:  
\begin{align}
    \hat{y}_{ij} = f_\theta(X_{ij}) = \argmax _{k \in \{0,..,9\}} P_\theta (y_{ij} = k |X_{ij}),
\end{align}

Once all the given images are predicted, the CP component finds a solution  $S \in \CSP (C_{rules},\{\hat{y}_{ij} \}^{given}) $ as visualised in Fig.~\ref{fig:NeuralNet1}.

From an inference standpoint, the above approach commits to the independent predictions made by the classifier and tries to use them as best as possible. 

\subsection{Hybrid1: reasoning over class probabilities}\label{section:ours1}

\begin{figure}
  \includegraphics[width=\linewidth]{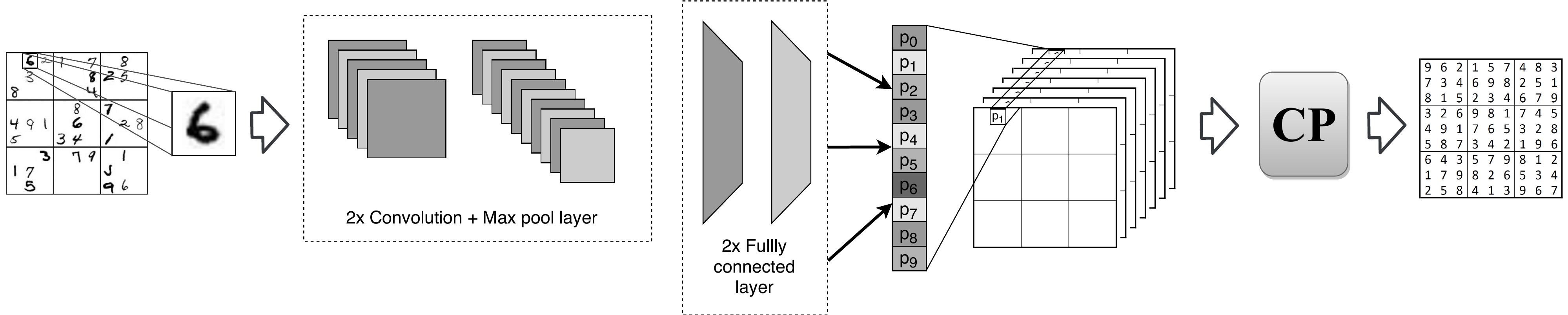}
  \caption{Architecture of class-probability reasoning approach}
  \label{fig:NeuralNet2}
\end{figure}

In this approach, we will use the same DNN architecture for digit classification as before. However, instead of using the hard labels from the DNN model, we will make use of the class probabilities of each of the given cells. 
Hence the outputs of the DNN, i.e., the inputs to the CP solver for each of the given cells, are $9$ probabilities -- one for each digit that can appear in a sudoku cell. The idea is to completely solve a sudoku grid by solving a $\COP$. See Fig~\ref{fig:NeuralNet2} for a visual representation of the architecture. 

Note that here, we make a joint inference over all the predictions, including their effect and relation to the a-priori empty cells. In the resulting solution, the digits of both given and non-given cells are obtained at once, while satisfying all the sudoku constraints.


First, the DNN is trained on images of single handwritten digits as before. After training, we store the DNN computed probabilities $P_\theta (y_{ij} = k |X_{ij})$ for each of the given $X_{ij}$.
We wish to make the CP solver reason (do \textit{inference}) over these probabilities directly, hence the sudoku problem formulation of Eq~\ref{eq:sudoku} needs to be modified to accommodate the probabilities. Instead of only satisfying the regular sudoku constraints of Eq~\ref{eq:sudoku}, we seek to find a solution which \emph{optimizes} the likelihood of the solution, given the probabilities obtained from the classifier.

More specifically, as each image is predicted on its own, we assume each to be an observation of an independent random variable, and hence the most likely solution is the one that maximizes the joint probability over the given images $\max \prod_{given (i,j)} \prod_{k \in \{1,..,9\}} (P_\theta (y_{ij} = k |X_{ij}))^{ \mathbbm{1}[s_{ij} = k]}$ for a solution $s$. We would like to find the most likely solution that also satisfies all constraints. After a log-transform, we can write the joint probability as a weighted sum objective function as follows:
\begin{align}
min  \sum_{\substack{(i,j)  \in \\ given }} \sum_{\substack{k \in \\ \{1,..,9\}}}  -log(P_\theta (y_{ij} = k |X_{ij}))*\mathbbm{1}[s_{ij} = k] \label{eq:ours1}
\end{align}
Treating $-log(P_{ij})$ as a $k$-dimensional vector, one can see that the inner sum could be formulated with a traditional \textit{element} constraint in a CP solver. 
We must emphasize that the log-likelihood is maximized only over the given cells and not for the whole grid due to the fact that we have the classifier provided probability vector only for these cells with given images.

Note that in this approach, the CP solver has to solve a more complex problem with larger domains for the given cells, and hence a larger search space. Contrary to the approach in section~ \ref{section:baseline} where the problem was a CSP, here the problem is a COP. 
The advantage of this approach is that it makes use of the constraint relationships of the sudoku problem. Moreover, it improves the prediction of the ML classifier by reasoning over these constraint relationships.
\subsection{Hybrid2: Higher-order knowledge exploitation}\label{section:ours2}
As mentioned before, a sudoku must have a unique solution for a set of givens. For traditional sudoku puzzles this is the case by construction, as otherwise, a human solver would be faced with having to choose among two or more options, rather than reasoning up to a full solution. 

In the approach of section~\ref{section:ours1}, we simply find one solution and treat that as \textit{the} solution, without verifying whether it is unique \textit{with respect to the set of givens}. When projecting the solution of the entire sudoku back to only the assignment to the `given' cells, e.g. those for which an image is given, then this assignment to the givens should have one and only one unique solution. If not, this assignment to the givens, and hence the entire sudoku solution, can not be the intended solution.

Therefore, we can use the (non) existence of a unique solution as an additional relational property that can steer the joint inference. The pseudo-code of this approach is shown in Algorithm~\ref{algo:ours2}. We start with finding the most likely solution $sol$ as in the \texttt{hybrid1} approach described in the previous section. We will write $\{sol_{ij}\}^{given}$ to represent the projected part of the solution, that is, only the part of the assignment of the cells with an image given.

Instead of \textit{counting all} solutions given $\{sol_{ij}\}^{given}$, it is sufficient (and computationally cheaper) to only check whether \textit{any }other solution exists. Hence, we will search for \textit{any} sudoku solution (line~\ref{alg:next}) that is different from the $sol$ solution that we already know exists (line~\ref{alg:tmpnogood}).

If there does not exist such other solution, i.e. the assignment is an empty set (line~\ref{alg:empty}), then the solution is unique and there is nothing more we can infer.
If there is another solution, we reject $\{sol_{ij}\}^{given}$ for not being unique. That is, we add a \textit{nogood} ensuring that no completion of $\{sol_{ij}\}^{given}$ will be found anymore (line~\ref{alg:nogood}), and repeat the procedure.

This use of a nogood, or a blocking clause, is common in solving such second-order logic problems. It can be seen as an instantiation of \textit{solution dominance}~\cite{DBLP:journals/corr/abs-1812-09207}.

\begin{algorithm}
\SetAlgoLined
$sol \gets \texttt{VizSudoku}(C_{rules}, \{X_{ij} \}^{given})$ \quad // as in hybrid1 \\

$C'_{rules} \gets C_{rules} \wedge \neg (V = sol))$ \quad \quad \quad \quad // temporarily forbid this solution \\ \label{alg:tmpnogood}
$sol' \gets \texttt{CSP}(C'_{rules}, \{sol_{ij}\}^{given})$ \quad \quad // check for other solutions having these givens \\ \label{alg:next}

\While{ $sol' \neq \emptyset $ \label{alg:empty}} 
{
  $C_{rules} \gets C_{rules} \wedge \neg (V^{given} = sol^{given})$ \quad // add nogood on givens \\  \label{alg:nogood}
  
  $sol \gets \texttt{VizSudoku}(C_{rules}, \{X_{ij} \}^{given})$ \quad \quad // as in hybrid1 \\

  $C'_{rules} \gets C_{rules} \wedge \neg (V = sol))$ \quad \quad \quad // temporarily forbid this solution \\
  $sol' \gets \texttt{CSP}(C'_{rules}, \{sol_{ij}\}^{given})$ \\
}
 \Return $sol$
 \caption{Higher-order COP of $ \texttt{VizSudoku}(C_{rules}, \{X_{ij} \}^{given})$ using a trained DNN $f_\theta(X)$}
 \label{algo:ours2}
\end{algorithm}

\section{Class probability calibration} \label{section:calibration}
In a machine learning context, \textit{calibration} is the process of modifying the predicted probabilities so that they match the expected distribution of probabilities for each class~\cite{guo2017calibration}. 
We will investigate the effect of calibration on our joint inference approach. Our method reasons over all 9 probability estimates $\{(P_\theta(y=1|X), \ldots, P_\theta(y=p|X)\}, pos\}$ and actively trades-off the probability of a prediction of one image to the prediction of another image in its objective function. Hence, it is not just a method of getting the top-predicted value right, but rather of getting all predicted probabilities correctly. Our reasoning approach hence assumes real (calibrated) probabilities and could be hampered by over- or under-confident class probability estimations.

In a multi-class setting, for a given handwritten digit a neural probabilistic classifier computes a vector $\Vec{z}$ containing raw scores for each class (i.\ e.\ a digit value), $\Vec{z_k}$ being the score assigned to class $k$. The SoftMax function is then applied to convert these raw scores into probabilities: 
$$\sigma_{\mathrm{SoftMax}}\left( \Vec{z}_{k}, \Vec{z}\right) = \frac{\exp \left(\Vec{z}_{k}\right)}{\sum_{i} \exp \left(\Vec{z}_{i}\right)}. $$ such that $P_\theta(y=k|X) = \sigma_{\mathrm{SoftMax}}\left( \Vec{z}_{k}, \Vec{z} \right) $ is the output of the neural network.

While this output is normalized across classes to sum up to 1, the values are not real probabilities. More specifically, it has been shown that especially neural networks tend to overestimate the probability that an item belongs to its maximum likelihood class \cite{guo2017calibration}.

Post-processing methods such as Platt scaling \cite{Platt99probabilisticoutputs} aim at calibrating the probabilistic output of a pre-trained classifier. Guo et al.~\cite{guo2017calibration} describe three variants of Platt scaling in the multi-class setting. In \emph{matrix scaling}, a weight matrix $\Vec{W}$ and a bias vector $\Vec{b}$ apply a linear transform to the input vector of the softmax layer $\Vec{z}_i$
such that the calibrated probabilities  become:
\begin{align}
\widetilde{P}_\theta(y_i=k|X_i)= \sigma_{\mathrm{SoftMax}}\left(\Vec{W_k} \Vec{z_k}+\Vec{b_k}, \Vec{W} \Vec{z}+\Vec{b}\right) 
\end{align} where $\Vec{W}$ and $\Vec{b}$ are parameters, learned by minimizing the Negative Log Likelihood loss on a validation set.
\emph{Vector scaling} applies the same linear transform, except that $\Vec{W}$ is a diagonal matrix, that is, only the diagonal is non-zero. Finally, \emph{Temperature scaling} considers a single scalar value $T$ to calibrate the probability such that:
\begin{align}
\widetilde{P}_\theta(y_i=k|X_i) = \sigma_{\mathrm{SoftMax}}\left( \dfrac{\Vec{z}_{k}}{T}, \dfrac{\Vec{z}}{T}\right)
\end{align}

To \textit{calibrate} the predictions, we train a model $f_{\theta,\Vec{W},\Vec{b}}(X)$ where $\{(\widetilde{P}_\theta(y=1|X), \ldots, \widetilde{P}_\theta(y=p|X)\}\}$ is calibrated on a validation set $\{X_i,y_i\}_{validation}$. More specifically, we will do calibration on top of a pre-trained neural network, so $\theta$ is pre-trained and the calibration learns the best $\Vec{W},\Vec{b}$.

We will evaluate whether better calibrated probabilities lead to better joint inference reasoning in the experiments.







\section{Experiments}

Numerical experiments were done on a subset of the Visual Sudoku Dataset used in \cite{wang2019satnet}. The subset contains 3000 sudoku boards whose givens are represented by MNIST digits. The average number of givens per sudoku grid is $36.2$.
Unless stated otherwise, the MNIST train data was split into $80\%-20\%$ train and validation set. 

The DNN architecture for the digit classification task is the LeNet architecture \cite{lecun1998gradient} which uses two convolutional layers followed by two fully connected layers. The network is trained for 10 epochs to minimize cross-entropy loss, and is optimized via Adam with a learning rate of $10^{-5}$. 
Once trained on the MNIST train data, we use the same model for both separate and hybrid approaches.
The neural network and CP model were implemented using PyTorch 1.3.0 \cite{paszke2017automatic} and OR-tools 7.4.7247 \cite{ortools}, respectively. All experiments were run on a laptop with 8 $\times$ Intel® Core™ i7-8565U CPU @ 1.80GHz and 16 Gb of RAM.


To test the performance of our proposed frameworks, we define the following evaluation measures:
\begin{description}[leftmargin=2\parindent,labelindent=\parindent]
\item{\texttt{img} accuracy} = percentage of givens correctly labeled by the classifier
\item{\texttt{cell} accuracy} = percentage of cells matching the true solution
\item{\texttt{grid} accuracy} = percentage of correctly solved sudokus. A sudoku is correctly solved if its true solution was found. That is, if \[s_1 \in \texttt{VizSudoku} (C_{rules},\{X_{ij}\}^{given}) \]
\[ \ s_2 \in \CSP(C_{rules}, \{y_{ij}\}^{given}) \implies s_1 \equiv s_2\]
\item{\texttt{failure rate grid}} = percentage of sudokus without a solution. A sudoku has no solution if $\texttt{VizSudoku} (C_{rules},\{X_{ij}\}^{given}) = \emptyset$
\end{description}
In the subsequent experiments, we denote as \texttt{baseline} the separated classification and reasoning approach, whereas we refer to our proposed approaches as \texttt{hybrid1} and \texttt{hybrid2}.

\subsection{Separate vs Hybrid Approaches}
First we compare the result of the three approaches described in section~\ref{section:methodology}.
As displayed on Table~\ref{tab:exp1}, the ability of the \texttt{baseline} approach to handle the image classification task with an accuracy of $94.75 \% $ translates to a meagre success rate of only $14.67\%$ at the level of sudoku grids correctly solved.
This is because the constraints relationships are not translated to the DNN model. As a consequence there is no way to ensure that the predictions would respect the constraints. Even a single mistake in predictions out of all the given images may result in an unsolvable puzzle. As an example, if one prediction error makes the same number appear twice in a row then the whole puzzle will be unsolvable even if the rest of the predictions are accurate. 



\begin{table}[t]
\setlength{\tabcolsep}{10pt}
\setlength{\belowcaptionskip}{-10pt}
\centering
\begin{tabular}{@{\extracolsep{\fill}}llrrrr@{}}
\toprule
              & \multicolumn{3}{c}{\textbf{accuracy}}                                                                    & \multicolumn{1}{c}{\textbf{failure rate}} & \multicolumn{1}{c}{\textbf{time}}        \\
              & \multicolumn{1}{c}{\textbf{img}} & \multicolumn{1}{c}{\textbf{cell}} & \multicolumn{1}{c}{\textbf{grid}} & \multicolumn{1}{c}{\textbf{grid}}         & \multicolumn{1}{c}{\textbf{average (s)}} \\ \midrule
baseline & 94.75\%                                 & 15.51\%                                  & 14.67\%                                  & 84.43\%                                  & 0.01                               \\
hybrid1    & 99.69\%                                 & 99.38\%                                  & 92.33\%                                  & 0\%                                   & 0.79                               \\
hybrid2    & 99.72\%                                 & 99.44\%                                  & 92.93\%                                  & 0\%                                   & 0.83                              \\ \bottomrule
\end{tabular}
\caption{Comparison of hybrid solving approaches}
\label{tab:exp1}
\end{table}

On the other hand the hybrid approaches do not consider the model predictions as final and by using the constraints relationships, \texttt{hybrid2}, for instance, brings the classifier to correctly label 5361 additional images. As a result we observed an increase in overall accuracy of the predictions. The advantage of our frameworks is more prominent from the \texttt{grid} perspective, where we can see that more than 92\% of the sudokus are now correctly solved. This is a huge improvement from the \texttt{baseline} approach which solves only 14.67\% of the grids. 

In terms of final performance \texttt{hybrid2} is more accurate as it exploits one more sudoku property; namely that sudoku must have a unique solution. By this mechanism we are able to further rectify more predictions and 18 additional puzzles are solved accurately. 

However, from a computational standpoint, our hybrid approaches solve a COP instead of a CSP in the pure reasoning case. Hence they are almost a 100 times more time consuming (only the average per sudoku is shown). The average computation time is slightly higher for \texttt{hybrid2} as we need to prove that predicted givens only have a unique solution, or optimize again with a forbidden assignments if that is not the case; this situation happens 18 times in our experiments.

\subsection{Reasoning Over Top-$k$ Probable Digits}
\begin{table}[b]
\setlength{\belowcaptionskip}{-10pt}
\centering
\begin{tabular}{@{}lrrrrrrrrr@{}}
\toprule
\textbf{} & \multicolumn{1}{l}{\textbf{rank-0}} & \multicolumn{1}{l}{\textbf{rank-1}} & \multicolumn{1}{l}{\textbf{rank-2}} & \multicolumn{1}{l}{\textbf{rank-3}} & \multicolumn{1}{l}{\textbf{rank-4}} & \multicolumn{1}{l}{\textbf{rank-5}} & \multicolumn{1}{l}{\textbf{rank-6}} & \multicolumn{1}{l}{\textbf{rank-7}} & \multicolumn{1}{l}{\textbf{rank-8}} \\ \midrule
hybrid1     & 94.85\%                            & 3.68\%                             & 0.93\%                             & 0.32\%                             & 0.12\%                             & 0.07\%                             & 0.02\%                             & 0.01\%                             & 0.01\%                             \\
hybrid2     & 94.84\%                            & 3.68\%                             & 0.92\%                             & 0.33\%                             & 0.12\%                             & 0.06\%                             & 0.02\%                             & 0.01\%                             & 0.01\%                             \\ \bottomrule
\end{tabular}
\caption{Rank distribution for cell values in correctly solved sudokus}
\label{tab:exp-rank1}
\end{table}

We are curious to know how the hybrid approaches outperform the separate approach. 
So we investigate when a digit is chosen by the hybrid approaches, how, on average, it is ranked by the ML classifier when ranking by probability. 

 Table~\ref{tab:exp-rank1} reveals, 
 among the instances where we find the correct solution, that the top-ranked value is chosen in most cases, with a quick decline in how often the other values are chosen. Remarkably, in 42 cases (i.e. $0.02 \% $ of predictions) \texttt{hybrid2} actually uses a digit which is ranked 8 or lower by the classifier.
 
From a combinatorial optimisation perspective, one can also consider that this allows to trade-off the size of the search space with the accuracy of the resulting solutions by only taking the $k$ highest probable digits into account and removing the others from the domains.
In this regard the experiment in the previous section considered two extremes: the baseline uses only the maximum probable digit, and the hybrid approaches use all $9$ digits.

Therefore, we investigate the effect of considering the top-k probability ranked digits on computational time and accuracy. Table~\ref{tab:exp-rank2} 
shows the effect of using only reasoning over the top-$k$ predicted values of the classifier:

\begin{table}[b]
\setlength{\tabcolsep}{10pt}
\setlength{\belowcaptionskip}{-10pt}
\centering
\begin{tabular}{@{}llrrrr@{}}
\toprule
              & \multicolumn{3}{c}{\textbf{accuracy}}                                                                    & \multicolumn{1}{c}{\textbf{failure rate}} & \multicolumn{1}{c}{\textbf{time}}        \\
              \textbf{top-$k$}
              & \multicolumn{1}{c}{\textbf{  img  }} & \multicolumn{1}{c}{\textbf{cell}} & \multicolumn{1}{c}{\textbf{grid}} & \multicolumn{1}{c}{\textbf{grid}}         & \multicolumn{1}{c}{\textbf{average (s)}} \\ \midrule
top-1 & 94.75\%                                 & 15.36\%                                  & 14.67\%                                  & 84.60\%                                  & 0.03                                 \\
top-2 & 96.15\%                                 & 63.63\%                                  & 55.43\%                                  & 34.20\%                                  & 0.03                                 \\
top-3 & 96.63\%                                 & 94.73\%                                  & 77.17\%                                  & 0.20\%                                   & 0.06                                 \\
top-4 & 98.78\%                                 & 98.04\%                                  & 86.33\%                                  & 0\%                                   & 0.12                                 \\
top-5 & 99.35\%                                 & 98.86\%                                  & 89.67\%                                  & 0\%                                   & 0.26                                 \\
top-6 & 99.57\%                                 & 99.21\%                                  & 91.60\%                                  & 0\%                                   & 0.38                                 \\
top-7 & 99.67\%                                 & 99.36\%                                  & 92.33\%                                  & 0\%                                   & 0.55                                 \\
top-8 & 99.69\%                                 & 99.40\%                                  & 92.63\%                                  & 0\%                                   & 0.66                                 \\
top-9 & 99.71\%                                 & 99.43\%                                  & 92.90\%                                  & 0\%                                   & 0.80                                 \\ \bottomrule
\end{tabular}
\caption{Rank experiment using hybrid2 for joint inference}
\label{tab:exp-rank2}
\end{table}

When considering top-1 to top-4 values, we see that the image accuracy steadily goes up as does the grid correctness, and grid failure reaches 0 for top-4. As we consider 4 or more digits, both \texttt{grid} and \texttt{image} values slowly increase, with the best results obtained using all possible values; which makes the difference for 8 sudoku instances when using \texttt{hybrid2}.

This shows that there is indeed a trade-off between computational time of the joint inference and accuracy of the result, with runtime performance gains possible at low accuracy cost if needed. 

\subsection{Classifier strength versus reasoning strength}
\begin{figure}[t]
\setlength{\belowcaptionskip}{-5pt}
\centering
\begin{minipage}{.3\textwidth}
  \centering
  \includegraphics[width=\linewidth]{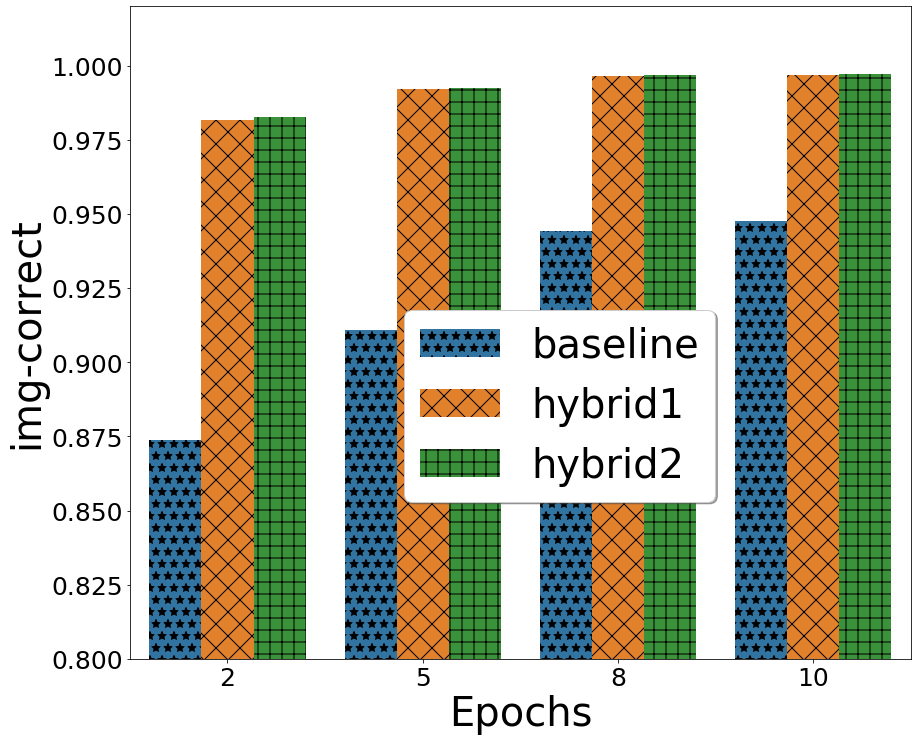}
  \subcaption{img-correct}
  \label{fig:test1}
\end{minipage}%
\begin{minipage}{.3\textwidth}
  \centering
  \includegraphics[width=\linewidth]{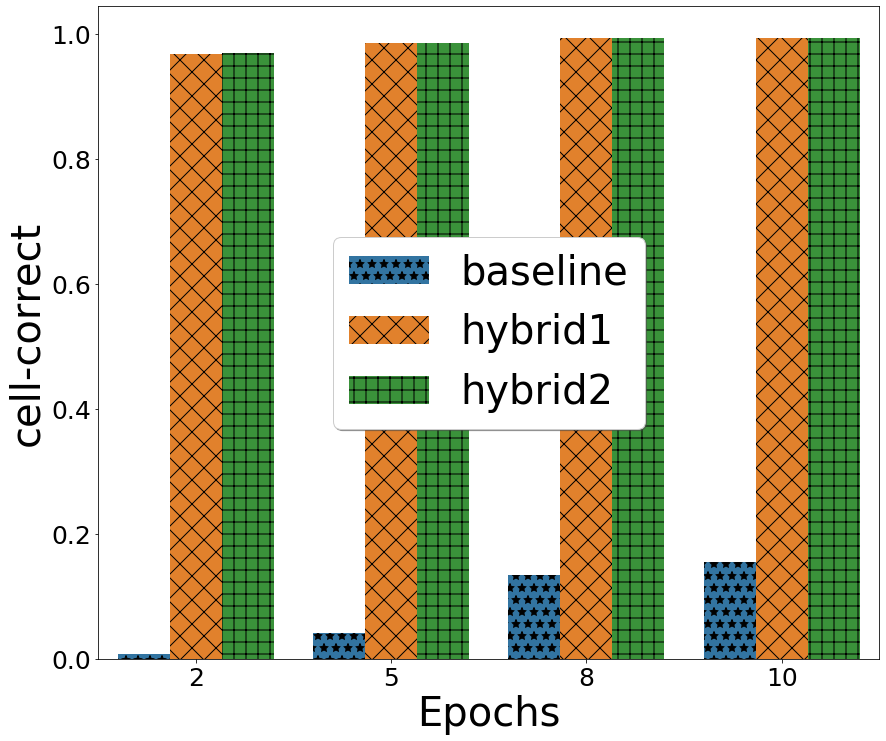}
  \subcaption{cell-correct}
  \label{fig:test2}
\end{minipage}
\begin{minipage}{.3\textwidth}
  \centering
  \includegraphics[width=\linewidth]{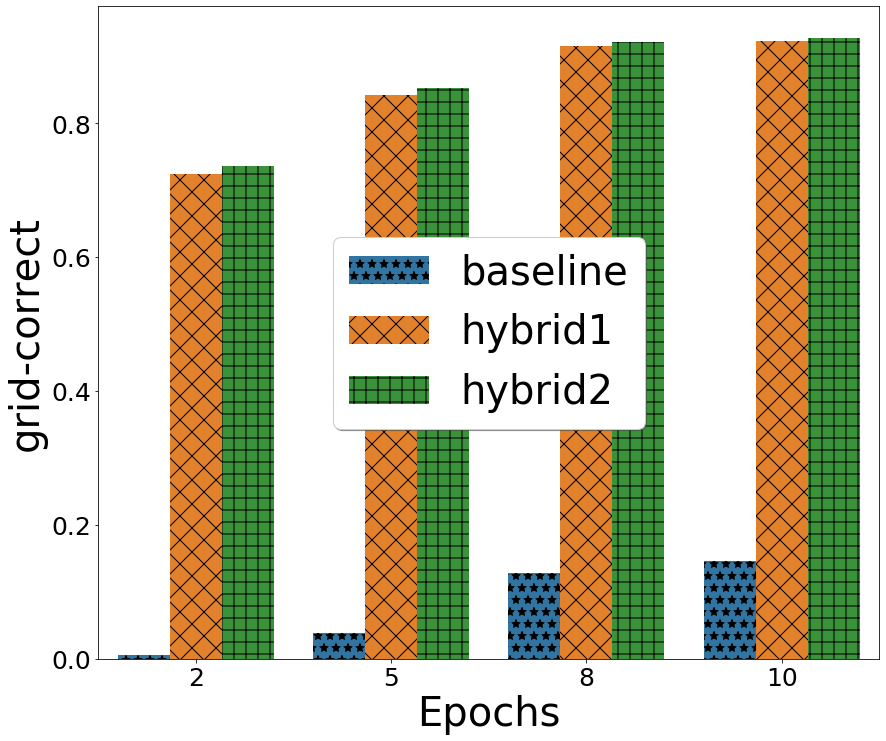}
  \subcaption{grid-correct}
  \label{fig:test3}
\end{minipage}%
\caption{Strength of hybrid with less accurate predictions \tias{This is impossible to interpret (for me)... I think per type (like Fig 3) would be a bit better, but if even that has too little difference in values then we should go back to tables...}}
\label{fig:classifier_strength}
\end{figure}

\begin{table}[t]
\centering
\setlength{\tabcolsep}{10pt}
\begin{tabular}{lrrrr}
\hline
         & \multicolumn{3}{c}{\textbf{accuracy}}                                                                    & \multicolumn{1}{c}{\textbf{failure rate}} \\
         & \multicolumn{1}{l}{\textbf{img}} & \multicolumn{1}{c}{\textbf{cell}} & \multicolumn{1}{c}{\textbf{grid}} & \multicolumn{1}{c}{\textbf{grid}}         \\ \hline
baseline & 99.384\%                          & 80.380\%                           & 80.100\%                           & 19.6\%                                   \\
hybrid1  & 99.984\%                         & 99.966\%                          & 99.500\%                          & 0\%                                   \\
hybrid2  & 99.986\%                          & 99.972\%                           & 99.600\%                           & 0\%                                       \\ \hline
\end{tabular}%

\caption{Comparison of separate and hybrid approach with a stronger classifier}
\label{tab:exp3strclf}
\end{table}
So far, we have used a fairly accurate model. We have also seen that joint inference by constraint solving could indeed correct many of the wrong predictions. In this experiment, we investigate the limits of this `correcting' power of the reasoning. That is, for increasingly worse predictive models, we compare the accuracy of the baseline with our hybrid approaches.

Results in Figure~\ref{fig:classifier_strength}  show that even after 2 epochs, with an accuracy of approximately 88\%, the reasoning is able to correct this to 98\%, i.e., a correction factor of 10\%. Hence, with weaker predictive models, the reasoning has even more potential for correcting.

Results on Table~\ref{tab:exp3strclf} show that this trend remains true even with a stronger classifier, obtained by considering a learning rate of $2\times 10^{-3}. $
In the stronger classifier case, \texttt{hybrid2} correctly classifies $654$ more images than the \texttt{baseline}.

Also noteworthy is that the average runtime goes up by a significant factor, e.g., it is 10 times slower as the predictions become less accurate. Further investigation shows that the predicted values are less skewed at lower accuracy levels, e.g., the softmax probabilities are more similar and hence the branch-and-bound search takes more time in finding and proving optimality.

\subsection{Effect of calibration}
            

\begin{figure}
    \setlength{\belowcaptionskip}{-10pt}
    \centering
    \includegraphics[width=0.8\linewidth]{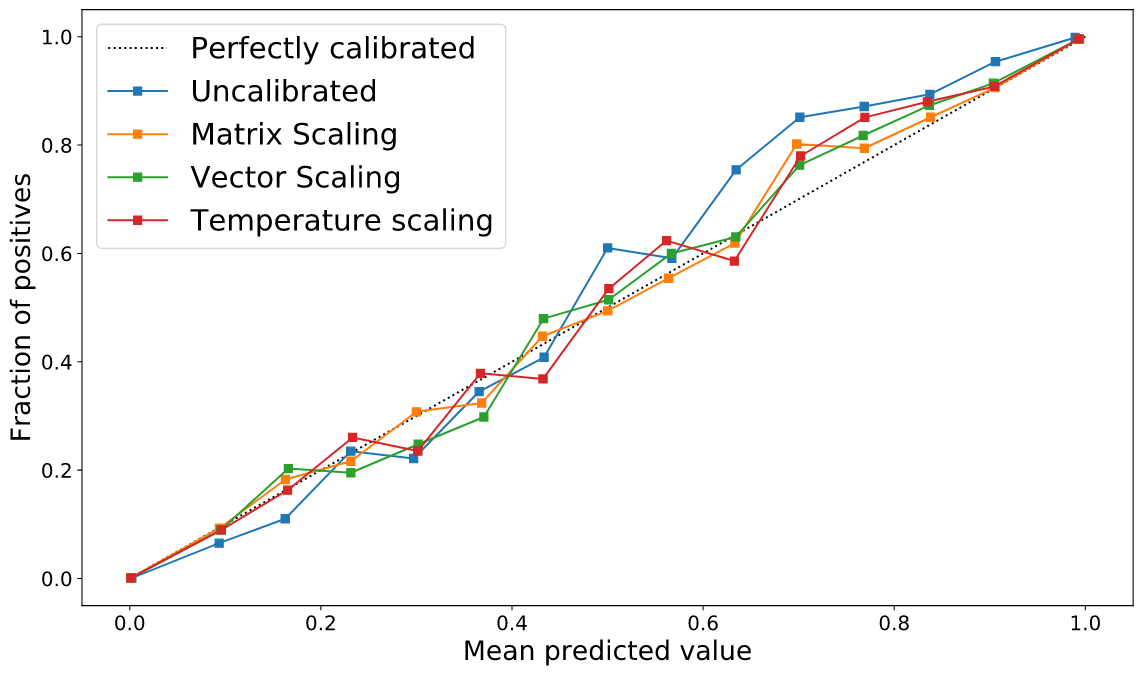}
    
    \caption{Calibration curve, mean of probabilities over 15 equally-sized intervals}
    \label{fig:calib_curve}
\end{figure}

\begin{table}[ht]
\setlength{\belowcaptionskip}{-20pt}
\centering
\begin{tabular}{@{}ll@{\hskip 0.2in}l@{\hskip 0.2in}l@{\hskip 0.2in}l@{}}
\toprule
          & \textbf{uncalibrated} & \textbf{Temp. scaling} & \textbf{Vector scaling} & \textbf{Matrix scaling} \\ \midrule
NLL       & 12.07              & 11.61                     & 11.38                & \textbf{10.12}       \\
test acc. & 96.75\%              & 96.75\%                     & 96.70\%                & \textbf{96.93\%}       \\ \bottomrule
\end{tabular}
\caption{NLL loss on validation set and test accuracy for Platt scaling variants}
\label{tab:calibdata}
\end{table}

\begin{figure}[b]
    \centering
    \includegraphics[width=\linewidth]{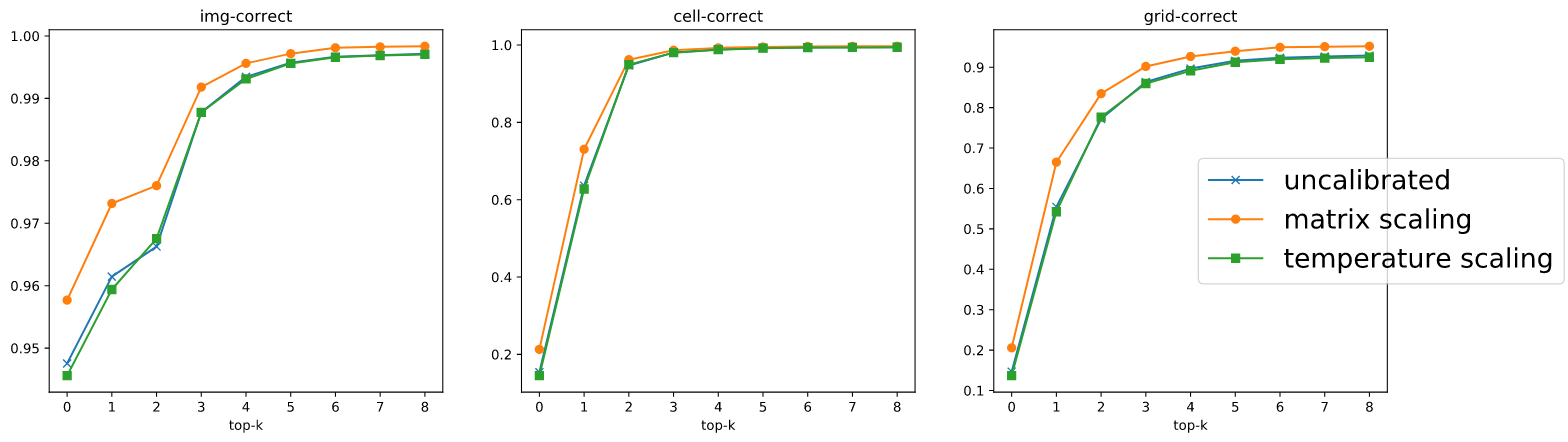}
    \caption{Performance measures for joint inference from calibrated classifier and comparison with uncalibrated counterpart}
    \label{fig:calibtop}
\end{figure}

As the joint inference reasons over the probabilities, we will investigate the effect of calibration on the reasoning. The first step towards that goal is to compare the different calibration methods we presented in section ~\ref{section:calibration}, namely \emph{Matrix scaling}, \emph{Vector scaling}, and \emph{Temperature scaling}. As described earlier, for each of these methods, calibration parameters are learned by minimizing the Negative Log Likelihood loss on the validation set (while remaining parameters of the network are fixed). 
Table~\ref{tab:calibdata} shows the validation NLL and the test accuracy before and after calibrating of the network. This table suggests that \emph{Matrix scaling} produces the most calibrated classifier. Figure~\ref{fig:calib_curve} shows how the classifier, although already quite well calibrated, is brought closer to a perfectly calibrated model. 

Figure~\ref{fig:calibtop} displays the effect of using a more calibrated model by running the top-k experiment with the \texttt{hybrid2} framework, with calibrated and uncalibrated classifiers.  It shows that calibration \emph{improves the accuracy} of our framework.
This is true when considering not only less accurate, but also more accurate, neural networks, as reasoning over all 9 probabilities leads a calibrated classifier used within the \texttt{hybrid2} framework to an \texttt{img} rate of $99.80\%$, an \texttt{accuracy cell} rate of $99.62\%$ and $94.30 \%$ of correctly solved grids.

\section{Conclusions}
In this paper we study a prototype application of hybrid prediction and constraint optimisation, namely the visual sudoku. Although deep neural networks have achieved unprecedented success in classification and reinforcement learning, they still fail at directly predicting the result of a combinatorial optimisation problem, due to the hard constraints and combinatorial optimisation aspect.

We propose a framework for solving challenging combinatorial problems like this, by adding a constraint programming layer on top of a neural network, which does joint inference over a set of predictions. We argue that reasoning over the actual predictions is limited as it ignores the probabilistic nature of the classification task, as confirmed by the experimental results. Instead, we can optimize the most likely joint solution over the classification probabilities which respects the hard constraints. Higher-order relations, such as that a solution must be unique, can also be taken into account to further improve the results.

Our proposed approach always finds a solution that satisfies the constraints, and \textit{corrects} the underlying neural network output up to 10\% in accuracy, for example transforming the output of a 94.8\% accurate classifier into a 99.7\% accurate joint inference classifier.

More broadly, we believe that this work is a notable path to incorporate domain-specific expertise in ML models. Practitioners often feel that they can help to make a ML model better by infusing their expertise into the model. However, incorporating such structured knowledge is often not feasible in a DNN setting. Our work proposes one way to impart human knowledge, namely on top of the neural network architecture and independent of the learning.

An interesting direction for future work is to look at differential classification+optimisation techniques, such as OptNet~\cite{amos2017optnet}, and investigate whether it is possible to train better models end-to-end for this kind of hard constrained problems. In this respect, there is also a link with probabilistic programming techniques, which often use knowledge compilation to embed (typically simpler) constraints in a satisfaction setting~\cite{manhaeve2018deepproblog}.
Finally, we are keen to apply this technique on applications involving
classification tasks, such as manhole maintenance~\cite{tulabandhula2013machine} and more.

\noindent \paragraph{\textbf{Acknowledgements.}}
This research received funding from the Flemish Government under the “Onderzoeksprogramma Artificiële Intelligentie (AI) Vlaanderen” programme.

\bibliographystyle{splncs04}       
{
\bibliography{refs}}

\end{document}